\pdfoutput=1
\documentclass{article}

\PassOptionsToPackage{numbers, compress}{natbib}
\usepackage[final]{neurips_2024}

\usepackage[utf8]{inputenc} %
\usepackage[T1]{fontenc}    %
\usepackage[colorlinks=true,
    linktoc=all]{hyperref}       %
\hypersetup{
    urlcolor=[rgb]{0,0,0.5},
    citecolor=[rgb]{0,0,0.5},
    linkcolor=[rgb]{0,0,0.5}
}
\usepackage{url}            %
\usepackage{booktabs}       %
\usepackage{amsfonts}       %
\usepackage{nicefrac}       %
\usepackage{microtype}      %
\usepackage{xcolor}         %
\usepackage{enumitem}
\usepackage{wrapfig}
\usepackage{soul}

\usepackage{graphicx}
\usepackage{nicefrac}
\usepackage{multicol}

\usepackage{algorithm}
\usepackage{algorithmicx}
\usepackage{algpseudocode}
\usepackage[ruled,vlined,noline,linesnumbered,algo2e]{algorithm2e}

\usepackage{style}
\usepackage{tabularx}
\usepackage{cleveref}
\usepackage{amsmath,amssymb}
\usepackage{url}
\usepackage{rotating}
\usepackage{tabularray}
\usepackage{booktabs}
\usepackage[normalem]{ulem}
\usepackage{subcaption}
\usepackage[abspage]{zref}

\DeclareMathOperator{\Tr}{Tr}

\usepackage{xcolor}

\newlength\myindent
\renewcommand{\Cref}[1]{\prettyref{#1}}
\newrefformat{eq}{Eq.~\ref{#1}}

\algdef{SE}[SUBALG]{Indent}{EndIndent}{}{\algorithmicend\ }%
\algtext*{Indent}
\algtext*{EndIndent}

\title{Personalized Federated Learning via Feature Distribution Adaptation}

\author{%
Connor J. McLaughlin,
Lili Su \\
Northeastern University, Boston, MA 02115\\
\texttt{\{mclaughlin.co,l.su\}@northeastern.edu}
}

\begin{document}
\setlength\intextsep{0pt}  %

\maketitle

\begin{abstract}

Federated learning (FL) is a distributed learning framework that leverages commonalities between distributed client datasets to train a global model. Under heterogeneous clients, however, FL can fail to produce stable training results. Personalized federated learning (PFL) seeks to address this by learning individual models tailored to each client. One approach is to decompose model training into shared representation learning and personalized classifier training. Nonetheless, previous works struggle to navigate the bias-variance trade-off in classifier learning, relying solely on limited local datasets or introducing costly techniques to improve generalization.
In this work, we frame representation learning as a generative modeling task, where representations are trained with a classifier based on the global feature distribution. We then propose an algorithm, pFedFDA, that efficiently generates personalized models by adapting global generative classifiers to their local feature distributions. 
Through extensive computer vision benchmarks, we demonstrate that our method can adjust to complex distribution shifts with significant improvements over current state-of-the-art in data-scarce settings. Our source code is available on GitHub\footnote{\url{https://github.com/cj-mclaughlin/pFedFDA}}.

\end{abstract}

\section{Introduction}
\label{sec:intro}
The success of deep learning models relies heavily on access to large, diverse, and comprehensive training data. However, communication constraints, user privacy concerns, and government regulations on centralized data collection often pose significant challenges to this requirement \cite{liu2020privacy, Pfeiffer_2023, Kaissis2021}. To address these issues, Federated Learning (FL) \cite{mcmahan2017communication} has gained considerable attention as a distributed learning framework, especially for its privacy-preserving properties and efficiency in training deep networks.

The FedAvg algorithm, introduced in the seminal work \cite{mcmahan2017communication}, remains one of the most widely adopted algorithms in FL applications \cite{majcherczyk2021flow, Wang_2022_CVPR, ramaswamy2019federated, yang2018applied, sheller2019multi, du2020federated}. It utilizes a parameter server to maintain a global model, trained through iterative rounds of distributed client local updates and server aggregation of client models. While effective under independent and identically distributed (\iid) client data, its performance deteriorates as client datasets become more heterogeneous (non-\iid). Data heterogeneity leads to the well-documented phenomenon of client drift \cite{karimireddy2020scaffold}, where distinct local objectives cause the model to diverge from the global optimum, resulting in slow convergence \cite{khaled2020tighter, li2019convergence} and suboptimal local client performance \cite{tan2022towards}. Despite extensive efforts \cite{li2020federated, karimireddy2020scaffold, wang2020tackling, deng2020adaptive} to enhance FedAvg for non-\iid clients, the use of a single global model remains too restrictive for many FL applications.

Personalized federated learning (PFL) has emerged as an alternative framework that produces separate models tailored to each client. 
The success of personalization techniques depends on balancing the bias introduced by using global knowledge that may not generalize to individual clients, and the variance inherent in learning from limited local datasets. Popular PFL techniques include regularized local objectives \cite{t2020personalized, li2021ditto}, local-global parameter interpolation \cite{deng2020adaptive}, meta-learning \cite{fallah2020personalized, jiang2019improving}, and representation learning \cite{oh2021fedbabu, collins2021exploiting, xu2023personalized, liang2020think}. While these techniques have shown significant improvements for clients under limited types of synthetic data heterogeneity (e.g., imbalanced partitioning of an otherwise \iid dataset), we find that current methods still struggle to navigate the bias-variance trade-off with the additional challenge of feature distribution shift and data scarcity, conditions commonly encountered in cross-device FL. 

\begin{figure}[t]
\begin{subfigure}{.5\textwidth}
    \centering
    \includegraphics[width=.95\linewidth]{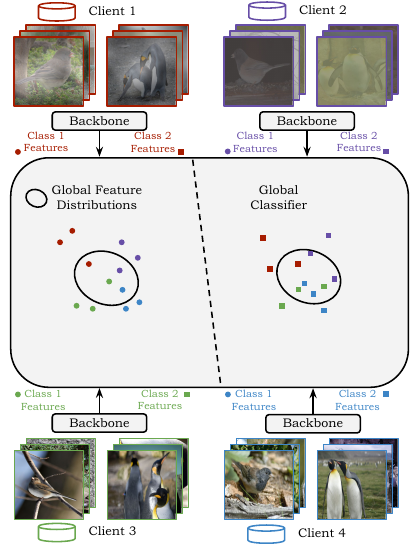}
    \caption{Global Representation Learning}
    \label{fig:global_representation_learning}
\end{subfigure}%
\begin{subfigure}{.5\textwidth}
    \centering
    \includegraphics[width=.95\linewidth]{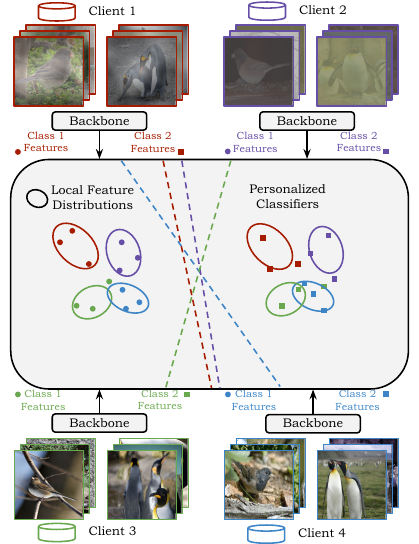}
    \caption{Local Distribution Adaptation}
    \label{fig:local_adaptation}
\end{subfigure}
\caption{\footnotesize
Overview of pFedFDA. (Left) Heterogeneous clients collaboratively train representation parameters under a generative classifier derived from a global estimate of class feature distributions. (Right) At test time, clients adapt the generative classifier to their feature distributions to obtain personalized classifiers.
}
\label{fig:overview}
\end{figure}

As such, we look to design a method capable of handling real-world distribution shifts, e.g., covariate shift caused by weather conditions or poor camera calibration, (see clients 1 and 2 in \prettyref{fig:overview}) with limited local datasets. To this end, we approach PFL through shared representation learning guided by a global, low-variance generative classifier. Specifically, we select a probability density $p$ with desirable statistical properties (e.g., one that admits an efficient Bayesian classifier) and iteratively estimate the global parameters of this distribution and representation layers to produce features from the estimated distribution (\prettyref{fig:global_representation_learning}).

To further navigate the bias-variance trade-off, we introduce a local-global interpolation method to adapt the global estimate to the distribution of each client. At inference time, clients use their adaptive local distribution estimate in a personalized Bayesian classifier (\prettyref{fig:local_adaptation}). 

{\bf Contributions.} We propose a novel 
{\bf P}ersonalized
{\bf Fed}erated
Learning method
based on
{\bf F}eature
{\bf D}istribution
{\bf A}daptation
({\bf pFedFDA}). We contextualize our algorithm using a class-conditional multivariate Gaussian model of the feature space in a variety of computer vision benchmarks. Our empirical evaluation demonstrates that our proposed method consistently improves average model accuracy in benchmarks with covariate shift or client data scarcity, obtaining over 6\% in multiple settings. At the same time, our method remains competitive with current state-of-the-art (often within 1\%) on more general benchmarks with more moderate data heterogeneity. 
To summarize, 
our contributions are three-fold:

\begin{itemize}[leftmargin=*,noitemsep,nolistsep] 
    \item A novel generative modeling perspective for federated representation learning is proposed to enable a new bias-variance trade-off for client classifier learning.  
    \item We propose a personalized federated learning method, pFedFDA, which leverages awareness of latent data distributions to guide representation learning and client personalization. 
    \item Extensive experiments on image classification datasets with varying levels of natural data heterogeneity and data availability demonstrate the advantages of pFedFDA in challenging settings. 
\end{itemize}

\section{Related Work}
\label{sec:intro}
{\bf Federated Learning with Non-\iid Data.}
Various studies have worked to understand and improve the ability of FL to serve heterogeneous clients. In non-\iid scenarios, the traditional FedAvg method ~\cite{mcmahan2017communication} is susceptible to client drift \cite{karimireddy2020scaffold}, resulting in slow convergence and poor local client accuracy \cite{li2019convergence, li2020federated}. 
To tackle this challenge,  \cite{li2020federated, acar2021federated, kim2022multi} proposed the use of regularized local objectives to reduce the bias on the global model after local training. Another approach focuses on rectifying the bias of local updates \cite{karimireddy2020scaffold, gao2022feddc} through techniques such as control variates. Other strategies include loss-balancing \cite{hsu2020federated, wang2021addressing, chen2021bridging}, knowledge distillation \cite{lin2020ensemble, zhu2021data}, prototype learning \cite{tan2022fedproto}, and contrastive learning \cite{li2021model}.
Despite promising results on non-\iid data, their reliance on a single global model poses limitations for highly heterogeneous clients \cite{kairouz2021advances}.

{\bf Personalized Federated Learning.}
In response to the limitations of a single global model, PFL seeks to overcome heterogeneity by learning models tailored to each client. In this framework, methods attempt to strike a balance between being flexible enough to fit the local distribution and relying on global knowledge to prevent over-fitting on small local datasets. Popular strategies include meta-learning an initialization for client adaptation \cite{jiang2019improving, fallah2020personalized}, multi-task learning with local model regularization \cite{t2020personalized, li2021ditto}, local and global model interpolation \cite{deng2020adaptive}, personalized model aggregation \cite{zhang2021personalized, zhang2022fedala}, client clustering \cite{sattler2019clustered, duan2021flexible}, and decoupled representation and classifier learning \cite{collins2021exploiting, liang2020think, oh2021fedbabu, xu2023personalized, chen2021bridging}. Our work focuses on this latter approach, in which the neural network is typically decomposed into the first $L-1$ layers used for feature extraction, and the final classification layer. 

Existing works in this category share feature extraction parameters between clients and rely on client classifiers for personalization. These approaches differ primarily in the acquisition of client classifiers during training, which influences representation learning. For example, FedRep \cite{collins2021exploiting} sequentially trains a strong local classifier while holding the representation fixed, then updates the representation under the fixed classifier. FedBABU \cite{oh2021fedbabu} proposes to use fixed dummy classifiers to align client objectives, only fine-tuning the classifier layers after the representation parameters have converged. Similarly, FedRoD \cite{chen2021bridging} aims to train a generic representation model and classifier in tandem via balanced softmax loss, later obtaining personalized classifiers through fine-tuning or hypernetworks. FedPAC \cite{xu2023personalized} adopts the learning algorithm of FedRep, but additionally regularizes the feature space to be similar across clients, before learning a personalized combination of classifiers across clients to improve generalization. However, this collaboration comes with an additional computational overhead that scales with the number of active clients. pFedGP \cite{achituve2021personalized} leverages a shared feature extractor as a kernel for client Gaussian processes. Although this approach offers improved sample efficiency, it comes at the cost of increased computational complexity and reliance on an inducing points set. 

In a similar spirit to our method, FedEM \cite{marfoq2021federated} estimates the latent data distribution of clients in parallel to training classification models. FedEM estimates each client data distribution as a mixture of latent distributions, where personalized models are a weighted average of mixture-specific models. Notably, this introduces a significant overhead in both communication and computation as separate models are trained for each mixture. In contrast, our work estimates the distribution of client features in parallel to training a global representation model.

\section{Problem Formulation}
\label{sec: preliminary}

{\bf FL System and Objective.}  
We consider an FL system where a parameter server coordinates with $M$ clients to train personalized models  %
$\theta_i$, $i=1,2, \cdots, M$. 
Each client $i$ has a local training dataset $\mathcal{D}_i=\{(x_i^j, y_i^j)\}_{j=1}^{n_i}$, where $x \in \reals^{m}$ and $y \in \{1, \cdots, C\}$.   
The model training objective in PFL is: %
\begin{align}
\min_{\theta_1, ..., \theta_M \in \calQ} ~ &f(\theta_1, ..., \theta_M) := \frac{1}{M} \sum_{i=1}^M F_i(\theta_i), 
\label{eq:pfl_objective}
\end{align}
where $\calQ$ is feasible set of model parameters, $F_i(\theta_i) = \mathbb{E}_{(x, y) \sim \mathcal{D}_i}{[L(\theta_i(x), y)]}$
is the empirical risk of dataset $\calD_i$, and $L$ is a loss function of the prediction errors (e.g., cross-entropy). The client population $M$ in FL can be large, resulting in partial client participation \cite{kairouz2021advances}. Let $q$ denote the participation rate, meaning that in each round, a client participates in model training with probability $q$. 

Following \cite{collins2021exploiting, xu2023personalized}, we approach this as a problem of global representation learning and local classification, in which each $\theta_i$ consists of a shared backbone,
$\phi$, responsible for extracting low-level features ($z \in \reals^{d} = \phi(x)$), and a local classifier,
$h_i$, for learning a client-specific mapping between features and labels. Considering this decomposition of parameters $\theta_i = (h_i\circ \phi)$, we can rewrite the original PFL objective as the following:
\begin{align}
\label{eq: objective rewritten}
\min_{\phi \in \Phi} \frac{1}{M}\sum_{i=1}^M \min_{h_i\in \calH} F_i(h_i\circ \phi),
\end{align} 
where $\Phi$ and $\calH$ are the feasible sets of neural network and classifier parameters, respectively. 

In our generative modeling framework, we consider $\calH$ to be the probability simplex over $\{1, \cdots, C\}$, and our algorithm uses approximations of the posterior distributions as classifiers $h_i$. However, for fair comparison with existing work (as well as other nice properties, discussed in \prettyref{sec:generative_model}), we select a generative model of the feature space such that $h_i$ can be represented with an equivalent linear layer.

\noindent 
{\bf Data Heterogeneity.}
The data distribution of each client $i$ is a joint distribution on $\calX\times \calY$, which can be written as $p_i(x,y)$, $p_i(y)p_i(x|y)$, or $p_i(x)p_i(y|x)$. Using the terminology of \cite{kairouz2021advances}, we refer to each case of data heterogeneity as follows: \textit{prior probability shift} ($p_i(y) \not =p_{i^{\prime}}(y)$), \textit{concept drift} ($p_i(x|y) \not =p_{i^{\prime}}(x|y)$), \textit{covariate shift} ($p_i(x) \not =p_{i^{\prime}}(x)$), and \textit{concept shift} ($p_i(y|x) \not =p_{i^{\prime}}(y|x)$). Furthermore, the local dataset volumes $\calD_i$ may have \textit{quantity skew}, i.e., $n_i\not=n_{i^{\prime}}$.

\section{Methodology}
\begin{wrapfigure}[21]{r}{.545\textwidth}
    \vspace*{-0.1\baselineskip}
    \centering
    \resizebox{0.95\linewidth}{!}{
    \renewcommand{\algorithmicrequire}{\textbf{Input:}}
\renewcommand{\algorithmicensure}{\textbf{Output:}}
\setlength{\textfloatsep}{1pt}%
\begin{algorithm2e}[H]
Server initializes feature extractor $\phi_g^{0}$ with random Gaussian weights. \label{line:phi_init}

Server and clients initialize feature distribution estimates ($\bm{\mu^0_g}, \Sigma^0_g$), ($\bm{\mu^0_i}, \Sigma^0_i$) as random spherical Gaussians. \label{line:gauss_init}

\For{each round $r=0, \cdots, R-1$}
{
    Send $\phi_g^{r}$, $\bm{\mu^r_g}, \Sigma^r_g$ to participating clients\\
    \For{each active client $i$}{
        $\phi_{i}^r \gets \phi_{g}^r$\\
        Train $\phi_{i}^{r}$ for $E$ epochs using loss \eqref{eq:global_objective} \label{line:representation_learning}\\
        Estimate local $\hat{\bm{\mu}}_i, \hat{\Sigma}_i$ via \eqref{eq:local-means} and \eqref{eq:local-cov} \label{line:gauss_estimate} \\
        Estimate interpolation parameter $\beta_i$ via \eqref{eq:beta} \label{line:beta} \\
        $\bm{\mu^r_i} \gets \beta_i \hat{\bm{\mu}} + (1-\beta_i) \bm{\mu}_g$ \\
        $\Sigma^r_i \gets \beta_i \hat{\Sigma} + (1-\beta_i) \Sigma_g$ \label{line:interpolation}\\
        Send $\phi_i^r, \bm{\mu^r_i}, \Sigma^r_i$ to the server \\
    }
    Update $\phi_g^{r}$, $\bm{\mu}_g^r$, $\Sigma_g^r$ with a weighted average of each client's $\phi_{i}^r$, $\bm{\mu}_i^r$, and $\Sigma_i^r$ \label{line:aggregation}
}
\caption{pFedFDA}
\label{algorithm:pfedfda_pseudocode}
\end{algorithm2e}}
\end{wrapfigure}
In this section, we introduce pFedFDA, a personalized federated learning method that utilizes a generative modeling approach to guide global representation learning and adapt to local client distributions. We present our method using a class-conditional Gaussian model of the feature space, with additional discussion of the selected probability density in \prettyref{sec:generative_model}. 

Algorithm \ref{algorithm:pfedfda_pseudocode} describes the workflow of pFedFDA. 

Our algorithm begins with a careful initialization of parameters for the feature extractor $\phi$, Gaussian means $\bm{\mu} = \{\mu^{c}\}_{c=1}^{C}$, and covariance $\Sigma$ (Lines \ref{line:phi_init}-\ref{line:gauss_init}). We initialize $\phi$ with established techniques (e.g., \cite{he2015delving}) such that the output features follow a Gaussian distribution with controlled variance. We similarly use a spherical Gaussian to ensure a stable initialization of the corresponding generative classifier (see \prettyref{sec:generative_model}). 

At the start of each FL round $r$, the server broadcasts the current $\phi_g^r, \bm{\mu}_g^r, \Sigma_g^r$ to each participating client. The local training of each client consists of two key components: (1) \textit{global representation learning}, in which clients train $\phi$ to maximize the likelihood of local features under the global feature distribution $\bm{\mu}_g^r, \Sigma_g^r$ (Line \ref{line:representation_learning}); (2) \textit{local distribution adaptation}, in which clients obtain robust estimates of their local feature distribution $\bm{\mu}_i^r, \Sigma_i^r$, using techniques for efficient low-sample Gaussian estimation (Line \ref{line:gauss_estimate}) and local-global parameter interpolation (Lines \ref{line:beta}-\ref{line:interpolation}). After local training, clients send their $\phi_i^r, \bm{\mu}_i^r, \Sigma_i^r$ to the parameter server for aggregation (Line \ref{line:aggregation}). 

In the following sections, we provide detailed explanations of each algorithmic component. In \prettyref{sec:generative_model} we discuss the benefits of a generative modeling framework and provide the justification for our selected class-conditional Gaussian model. We outline how the resulting generative classifier can be used to guide representation learning in \prettyref{sec:representation_learning} and describe how we obtain personalized generative classifiers in \prettyref{sec:local_adaptation}. 

\subsection{Generative Model of Feature Distributions}
\label{sec:generative_model}
\textbf{Motivation for Generative Classifiers.} A central theme in FL is exploiting inter-client knowledge to train more generalizable models than any client could attain using only their local dataset. This presents an important bias-variance trade-off, as incorporating global knowledge naively can introduce significant bias. %
Fortunately, under a generative modeling approach, this bias can be naturally handled, enabling efficient inter-client collaboration.

First note that local class priors $p_i(y)$ can be approximated with local counts: $p_i(y=c) \approx \frac{n_i^c}{\sum_{c' \in C} n_i^{c'}} := \pi_i^c$, where $n_i^c$ is the number of local samples whose labels are $c$. 
    This leaves the primary source of bias to the mismatch between local and global feature distributions $p_g(z|y)$ and $p_i(z|y)$. Crucially, it turns out that this bias is controllable due to the dependence of $z$ on global representation parameters $\phi$. Consequentially, we propose to minimize this bias through our classification objective, which we discuss further in \prettyref{sec:representation_learning}.

\textbf{Class-Conditional Gaussian Model.} In this work we approximate the distribution of latent representations using a class-conditional Gaussian with tied covariance, i.e., $p_i(z|y=c) = \calN(z|\mu_i^{c}, \Sigma_i)$. We show the resulting generative classifier under this model in \prettyref{eq:lda}. Note that it has a closed form and results in a decision boundary that is linear in $z$. I.e., if we know the underlying local feature distribution mean and covariance, we can efficiently compute the optimal header parameters $h_i$ for the inner objective in \prettyref{eq: objective rewritten}. 

In addition to the convenient form of the Bayes classifier, we select this distribution as the Gaussianity of latent representations is likely to hold in practice. Notably, by adopting the common technique of Gaussian weight initialization (e.g., \cite{he2015delving}), the resulting feature space is highly Gaussian at the start of training. It has also been observed that the standard supervised training of neural networks with cross-entropy objectives results in a feature space that is well approximated by a class-conditional Gaussian distribution \cite{lee2018simple}, i.e., the corresponding generative classifier \prettyref{eq:lda} has equal accuracy to the learned discriminative classifier. We provide a further discussion of this modeling assumption in \prettyref{appendix:limitations}.
\begin{align}
    p(y=c|z)      &= \frac{\calN(z|\mu^c, \Sigma)p(y=c)}{\sum_{c' \in \mathcal{C}} \calN(z|\mu^{c'}, \Sigma)p(y=c')}, \\
    \log{p(y=c|z)} &\propto z^{\top}\Sigma^{-1}\mu^{c} - \frac{1}{2}(\mu^{c})^{\top}\Sigma^{-1}\mu^{c} + \log{p(y=c)}. 
\label{eq:lda}
\end{align}
\vspace{-\baselineskip}

\subsection{Global Representation Learning}
\label{sec:representation_learning}

Next, we describe our process for training the shared feature extractor $\phi$. Similar to existing works \cite{collins2021exploiting, xu2023personalized}, our local training consists of training $\phi$ via gradient descent to minimize the cross-entropy loss of predictions from fixed client classifiers. We obtain our client classifiers through \prettyref{eq:lda}, using global estimates of $\bm{\mu}_g, \Sigma_g$ and local estimated priors $\pi_i$. For computational efficiency, we avoid inverting the covariance matrix by estimating $\Sigma^{-1}\mu^{c}$ with the least-squares solution $w = \min_{w'}\| \Sigma w' - \mu^c \|.$ 

The loss of client $i$ for an individual training sample ($x, y$) is provided in \prettyref{eq:global_objective}.
\begin{align}
    L(x, y; \phi, \bm{\mu}, \Sigma, \pi) &= \sum_{c=1}^{C} y^{c} \log{p(y^c|\phi(x), \mu^{c}, \Sigma, \pi)}.
\label{eq:global_objective}
\end{align}
\vspace{-\baselineskip}

Note that for a spherical Gaussian $\Sigma=\identity$ and uniform prior $\pi$, we recover a nearest-mean classifier under Euclidean distance. This resembles the established global prototype regularization \cite{tan2022fedproto}, which minimizes the Euclidean distance of features from their corresponding global class prototypes. Notably, FedPAC \cite{xu2023personalized} uses this prototype loss to align client features. However, this implicitly assumes that all feature dimensions have equal variance, and additionally requires a hyperparameter $\lambda$ to balance the amount of regularization with the primary objective. In contrast, our generative classifier naturally aligns the distribution of client features by training $\phi$ with our global generative classifier.

\subsection{Local Distribution Adaptation}

\label{sec:local_adaptation}

\textbf{Local Estimation.} 
A key component of pFedFDA is the estimation of local feature distribution parameters, used both for model personalization and for updating the global distribution for representation learning. 

Given a set of $n$ extracted features $Z$ with $n^c$ examples per class $c$, a maximum likelihood estimate of the class means and an unbiased estimator of the covariance, respectively, are given by:  
\begin{minipage}[h]{.45\textwidth}
\begin{align}
    \hat{\mu}^{c} &= \frac{1}{n^{c}} \sum_{j=1}^{n} \indc{{y_j = c}} z_j
    \label{eq:local-means}
\end{align}
\end{minipage}%
\hspace{1em}
\begin{minipage}[h]{.45\textwidth}
\vspace*{0pt}
\begin{align}
    \hat{\Sigma} &= \frac{1}{n-1} \bar{Z}^{\top}\bar{Z},
\label{eq:local-cov}
\end{align}
\end{minipage}

where, with slight abuse of notation, $\bar{Z}\in \reals^{n\times d}$ denotes the matrix of centered features with rows corresponding to each original feature $z_j$ centered by their respective means, i.e.,
\begin{align}
    \bar{z}_j &= z_j-\sum_{c \in C} \indc{{y_j = c}} \hat{\mu}^{c}. 
\label{eq:centered_features}
\end{align}
Estimators \prettyref{eq:local-means} and \prettyref{eq:local-cov} may be noisy on clients with limited local data. To illustrate this, consider the common practical scenario where $n_i \ll d$. The feature covariance matrix $\Sigma_i$ at client $i$ will be degenerate; in fact, it will have a multitude of zero eigenvalues.
In these cases, we can add a small diagonal $\epsilon{}\identity$ to $\Sigma$, and replace the non-positive-definite matrices with the nearest positive definite matrix with identical variance. This can be efficiently computed by clipping eigenvalues in the corresponding correlation matrix and followed by converting it back to a covariance matrix with normalization to maintain the initial variance. We refer readers to \cite{guo2007regularized} for a review of low-sample covariance estimation. 

\noindent 
\textbf{Local-Global Interpolation.}
We introduce this fusion because even with the aforementioned correction to ill-defined covariances, the variance of the local estimates remains highly noisy, indicating the necessity of leveraging global knowledge. It is essential to consider that in the presence of data heterogeneity, clients with differing local data distributions and dataset sizes have varying requirements for global knowledge. 

For our Gaussian parameters $\bm{\mu}, \Sigma$, we consider the introduction of global knowledge through a personalized interpolation between local and global estimates, which can be viewed as a form of prior. We provide an analysis of the high-probability bound on estimation error for an interpolated mean estimate in simple settings in Theorem \ref{Theorem:bias-variance}. The full derivation is deferred to \prettyref{appendix:proof}.

 \begin{theorem}[Bias-Variance Trade-Off]
    Let $C=1$. 
    Define $\mu_i$ as the sample mean of client $i$'s local features $\mu_i := \frac{1}{n_i}\sum_{j=1}^{n_i} z_{i}^j$, and $\mu_g$ as the global sample mean using all $N$ samples across $M$ clients: $\mu_g := \frac{1}{N}\sum_{i=1}^{M}\sum_{j=1}^{n_i} z_i^j$. Assume client features are independent and distributed as $z_i \sim \calN(\theta_i, \Sigma_i)$, with true global feature distribution $\calN(\theta_g, \Sigma_g)$. We consider the use of global knowledge at client $i$ through an interpolated estimate: $\hat{\mu}_i := \beta\mu_i + (1-\beta) \mu_g$, where $\beta \in [0, 1]$. 
    For any $\delta\in (0,1)$, 
    with probability at least $1 - \delta$, it holds that
    \begin{align*}
    \norm{\hat{\mu}_i - \theta_i}^2 
     &\le (1-\beta)^2 \norm{\theta_g - \theta_i}^2 \\
    & ~~~
    + \qth{1 + 4\pth{\sqrt{\frac{\log 1/\delta}{c}}+\frac{\log 1/\delta}{c}}}
    \pth{
    \frac{2\beta}{n_i}
    \Tr(\Sigma_i) + 
    \frac{(1-\beta)^2}{N}\Tr(\Sigma_g)},
    \end{align*}
    where $c>0$ is an absolute constant. 
    \label{Theorem:bias-variance}
\end{theorem}

Intuitively, the estimation error and optimal $\beta$ depend on the bias introduced by using global knowledge $\norm{\theta_g - \theta_i}^2$, the variance of local and global features, and the respective data volumes. 

We formulate this as an optimization problem, in which clients estimate interpolation coefficients $\beta_i$ to combine local and global estimates of ($\bm{\mu}, \Sigma$) with minimal $k$-fold validation loss:
\begin{align}
    \beta_i \in \min_{0 \leq \beta \leq 1} \hspace{0.5em} \frac{1}{k} \sum_{k} \sum_{(x, y) \in \mathcal{D}_k} L(x, y, \phi, \beta' \hat{\bm{\mu}}_k + (1-\beta') \bm{\mu}_g, \beta' \hat{\Sigma}_k + (1-\beta') \Sigma_g, \pi_i),
\label{eq:beta}
\end{align}
where $\mathcal{D}_k$ is the dataset consisting of the validation samples for the $k$-th fold, and ($\hat{\bm{\mu}}_k, \hat{\Sigma}_k)$ are the local distribution estimates \prettyref{eq:local-means} and \prettyref{eq:local-cov} estimated using the training samples from the $k$-th fold. In our experiments, we avoid additional forward passes on the local dataset by preemptively storing the feature-label pairs obtained over the latest round of training.

We solve \prettyref{eq:beta} using off-the-shelf quasi-newton methods (e.g., L-BFGS-B). We additionally explore using separate $\beta$ terms for the means and covariance (\prettyref{results:ablation}) and recommend the use of a single $\beta$ term for most applications. 

After obtaining $\beta$, we set our local estimates of $\bm{\mu}_i, \Sigma_i$ to their interpolated versions. These estimates are then sent to the server for aggregation. Notably, the server update rule can be viewed as a moving average \cite{zhang2015deep} between the previous round estimate and the client average scaled by $\beta$, reducing the influence of local noise in the global distribution estimate. At test time, clients use their local distribution estimates for inference through the classification rule in \prettyref{eq:lda}.

\vspace*{-\baselineskip}
\section{Experiments}
\vspace*{-.5\baselineskip}
\subsection{Experimental Setup}
{\bf Datasets, Tasks, and Models:} We consider image classification tasks and evaluate our method on four popular datasets. The EMNIST \cite{cohen2017emnist} dataset is for 62-class handwriting image classification. The CIFAR10/CIFAR100 \cite{cifar} datasets are for 10 and 10-class color image classification. The TinyImageNet \cite{le2015tiny} dataset is for 200-class natural image classification. For EMNIST and CIFAR10/100 datasets, we adopt the 4-layer and 5-layer CNNs used in \cite{xu2023personalized}. On the larger TinyImageNet dataset, we use the ResNet18 \cite{he2016deep} architecture. Notably, the feature dimension $d$ for EMNIST/CIFAR CNNs is 128, and 512 for ResNet. We provide additional details in \prettyref{appendix:datasets}.

\noindent
{\bf Clients and Dataset Partitions:} 
The EMNIST dataset has inherent covariate shifts due to the individual styles of each writer. We partition the dataset by writer following \cite{deng2020adaptive}, and train with $M=1000$ total clients (writers), participating with rate $q=0.03$. 
On CIFAR and TinyImageNet datasets, we simulate prior probability shift and quantity skew by partitioning the dataset according to a Dirichlet distribution with parameters $\alpha \in (0.1, 0.5)$, where lower $\alpha$ indicates higher levels of heterogeneity. On these datasets, we use $M=100$ clients with participation rate $q=0.3$. 
Additional details of the partitioning strategy are provided in \prettyref{appendix:partition}.

We split each client's data partition 80-20\% between training and testing. 

\noindent 
{\bf Covariate Shift and Data Scarcity:}
We introduce two modifications to client partitions to simulate the challenges of real-world cross-device FL. We first consider common sources of input noise for natural images, which may result from the qualities of the measuring devices (e.g., camera calibration, lens blur) or environmental factors (e.g., weather, lighting). To simulate this, we select ten image corruptions at five levels of severity defined in  \cite{hendrycks2019benchmarking}, and corrupt the training and testing samples of the first 50 clients in CIFAR10/100 with unique corruption-severity pairs. We leave the remaining 50 client datasets unchanged. We refer to these datasets with natural covariate shifts as CIFAR10-S/CIFAR100-S and detail the specific corruptions in \prettyref{appendix:cifar-s}.

Second, we perform uniform subsampling of client training sets, leaving them with (75\%, 50\%, or 25\%) of their original samples. These low-sample settings are more realistic for cross-device FL, where clients rely more on knowledge sharing.

\noindent
{\bf Baselines and Metrics:}
We compare pFedFDA to the following baselines: Local, in which each client trains its model in isolation; FedAvg \cite{mcmahan2017communication} and FedAvg with fine-tuning (FedAvgFT); APFL \cite{deng2020adaptive}; Ditto \cite{li2021ditto}; pFedMe \cite{t2020personalized}; FedRoD \cite{chen2021bridging}; FedBABU \cite{oh2021fedbabu}; FedPAC \cite{xu2023personalized}; FedRep \cite{collins2021exploiting}; and LG-FedAvg \cite{liang2020think}. We report the average and standard deviation of client test accuracies. 

\noindent 
{\bf Model Training:}
We train all algorithms with mini-batch SGD for $E=5$ local epochs and $R=200$ global rounds. We apply no data augmentation besides normalization into the range $[-1, 1]$. For pFedFDA, we use $k=2$ cross-validation folds to estimate a single $\beta_i$ term for each client. Additional training details and hyperparameters for each baseline method are provided in \prettyref{appendix:settings}.

\subsection{Numerical Results}
{\bf Performance under covariate shift and data scarcity.}
We first present our evaluation under natural client covariate shift with varying data scarcity in \prettyref{table:shift-scarce}. In all experiments, pFedFDA outperforms the other methods in test accuracy, demonstrating the effectiveness of our method in adapting to heterogeneous client distributions. Additionally, pFedFDA has an increasing benefit relative to other methods in data-scarce settings: on CIFAR10, we improve 4.2\% over the second-best method with 100\% of training samples and 6.9\% with 25\%. On CIFAR100, the same improvements range from 0.1\% to 6.5\%. This indicates the success of our method in navigating the bias-variance trade-off. 

\begin{table}[t]
\centering
\captionsetup{font=footnotesize}
\footnotesize
\caption{Average (standard deviation) test accuracy on CIFAR10/100-S for varying proportions of training data. }
\label{table:shift-scarce}
\setlength{\tabcolsep}{5pt}
\begin{tabular}{@{}lcccccccc@{}}
\toprule
Dataset             & \multicolumn{4}{c}{CIFAR10-S Dir(0.5)} & \multicolumn{4}{c}{CIFAR100-S Dir(0.5)} \\ \cmidrule(lr){2-5} \cmidrule(lr){6-9}
\% Samples          & 100     & 75     & 50     & 25    & 100     & 75     & 50     & 25     \\ \midrule
Local Only          & .586(.12) & .476(.16)              & .461(.15)              & .435(.14)              & .157(.05)               & .136(.05)              & .123(.04)              & .093(.04) \\ \midrule
FedAvg              & .464(.13) & .410(.19)              & .389(.17)              & .321(.14)              & .233(.06)               & .212(.06)              & .187(.05)              & .114(.04) \\
FedAvgFT           & \underline{.682(.10)} & .579(.19)              & .561(.17)              & .526(.16)              & .302(.06)               & .273(.05)              & .241(.06)              & .160(.05) \\ \midrule
APFL                & .611(.12) & .520(.17)              & .508(.16)              & .504(.16)              & .164(.05)               & .148(.04)              & .131(.05)              & .105(.04) \\
Ditto               & .668(.10) & .578(.18)              & .558(.17)              & .527(.16)              & .295(.05)               & .274(.06)              & .239(.05)              & .141(.05) \\
FedBABU             & .602(.12) & .522(.17)              & .495(.16)              & .467(.15)              & .187(.05)               & .170(.05)              & .148(.05)              & .107(.04) \\
FedPAC              & .679(.09) & \underline{.642(.19)}              & \underline{.594(.16)}              & \underline{.533(.18)}              & \underline{.360(.07)}               & \underline{.330(.07)}              & \underline{.283(.07)}              & \underline{.162(.05)} \\
FedRep              & .612(.10) & .541(.17)              & .510(.16)              & .486(.16)              & .176(.05)               & .158(.05)              & .131(.04)              & .100(.04) \\
FedRoD              & .655(.11) & .554(.18)              & .537(.18)              & .499(.14)              & .218(.05)               & .186(.05)              & .150(.04)              & .115(.04) \\
LG-FedAvg           & .584(.13) & .483(.16)              & .466(.15)              & .433(.14)              & .166(.05)               & .153(.05)              & .127(.05)              & .091(.04) \\
pFedMe              & .679(.10) & .583(.18)              & .549(.17)              & .523(.16)              & .289(.06)               & .268(.06)              & .237(.06)              & .153(.05) \\ \midrule
pFedFDA      & \textbf{.724(.09)} & \textbf{.706(.10)}              & \textbf{.661(.11)}              & \textbf{.595(.12)}              & \textbf{.361(.08)}               & \textbf{.342(.08)}              & \textbf{.326(.08)}              & \textbf{.227(.07)} \\ \bottomrule
\end{tabular}
\end{table}

\vspace{0.5em}
{\bf Evaluation in more moderate scenarios.}
Our evaluation of all four datasets in the traditional setting (no added covariate shift, full training data) is presented in \prettyref{table:general-results}. We note that: (1) our method is still competitive, always ranking within the top 3 methods, and (2) the gap between top methods is smaller than in the previous experimental setting. For example, on EMNIST/CIFAR10, we see that FedAvgFT, FedPAC, and pFedFDA are within $\sim$1\% accuracy. We observe larger performance gaps for CIFAR100, with FedPAC and pFedFDA having the best results.

\begin{table}[t]
\centering
\captionsetup{font=footnotesize}
\footnotesize
\caption{Average (standard deviation) test accuracy on multiple datasets.}\begin{tabular}{@{}lccccccc@{}}
\toprule
Dataset        & EMNIST  & \multicolumn{2}{c}{CIFAR10} & \multicolumn{2}{c}{CIFAR100} & \multicolumn{2}{c}{TinyImageNet} \\ \cmidrule(lr){2-2}\cmidrule(lr){3-4}\cmidrule(lr){5-6}\cmidrule(l){7-8}
Partition      & Writers & Dir(0.1)     & Dir(0.5)     & Dir(0.1)      & Dir(0.5)     & Dir(0.1)        & Dir(0.5)       \\ \midrule
Local & 
.242(.23) & .865(.13) & .585(.13) & .368(.09) & .150(.05) & .270(.07) & .099(.03) \\
\midrule
FedAvg & .790(.14) & .545(.12) & .625(.07) & .245(.06) & .252(.05) & .155(.04) & .150(.04) \\
FedAvgFT & \underline{.844(.10)} & .902(.10) & .742(.08) & .499(.09) & .314(.06) & .384(.07) & .213(.04) \\
\midrule
APFL & .841(.10) & .882(.11) & .656(.11) & .388(.09) & .169(.05) & .350(.09) & .177(.05) \\
Ditto & .843(.10) & .898(.10) & .736(.08) & .504(.08) & .308(.06) & \underline{.386(.07)} & .211(.04) \\
FedBABU & .728(.13) & .887(.11) & .678(.11) & .395(.09) & .193(.04) & .365(.07) & .179(.04) \\
FedPAC & \textbf{.856(.09)} & \textbf{.908(.09)} & \textbf{.767(.07)} & \textbf{.560(.08)} & \underline{.378(.06)} & .366(.07) & .180(.04) \\
FedRep & .735(.12) & .889(.10) & .668(.10) & .398(.09) & .182(.05) & .359(.07) & .145(.04) \\
FedRoD & .747(.15) & .885(.11) & .713(.09) & .424(.08) & .224(.05) & .382(.07) & .209(.05) \\
LG-FedAvg & .666(.13) & .866(.13) & .599(.12) & .381(.09) & .162(.05) & .280(.07) & .105(.03) \\
pFedMe & .842(.10) & .900(.10) & .740(.09) & .493(.08) & .311(.06) & \textbf{.388(.07)} & \textbf{.218(.04)} \\ \midrule
pFedFDA      & \underline{.844(.10)} & \underline{.902(.09)}              & \underline{.763(.07)}              & \underline{.523(.08)}              & \textbf{.385(.07)}                          & .384(.07)              & \underline{.214(.04)}              \\
\bottomrule
\end{tabular}
\label{table:general-results}
\end{table}

{\bf Results under extreme data scarcity.}
We present additional results at the limits of data scarcity on CIFAR10/100 datasets in \prettyref{table:extreme}, where we assign a single mini-batch (50) of training examples to each client. Notably, even as $n_i \ll d$, which poses a challenge to local covariance estimation, pFedFDA clients obtain the best test accuracy, indicating the robustness of our local-global adaptation.

\begin{table}[t]
\centering
\setlength{\tabcolsep}{4pt}
\captionsetup{font=footnotesize}
\footnotesize
\caption{Results under extreme data scarcity on CIFAR10/CIFAR100 Dir(0.5).}
\label{table:extreme}
\begin{tabular}{@{}lccccccccc@{}}
\toprule
        & FedAvgFT  & Ditto &  APFL & FedBABU & pFedMe & FedRoD & FedPAC & pFedFDA  \\ \midrule
CIFAR10 & .692(.17) & .708(.17) & .698(.17) & .684(.19) & .683(.16) & .631(.19) & \underline{.710(.16)} & \textbf{.725(.16)} \\
CIFAR100 & .324(.14) & .308(.14) & .405(.15) & .369(.15) & .305(.14) & .306(.14) & \underline{.407(.15)} & \textbf{.499(.16)}\\ \bottomrule
\end{tabular}
\end{table}

{\bf Generalization to new clients.}
We further analyze the ability of our generative classifiers to generalize on clients unseen at training time. To simulate this setting, we first train the server model model using half of the client population. We then evaluate each method on the set of clients not encountered throughout training, using their original input data, as well as their dataset transformed using each corruption from CIFAR-S. Further benchmark details, including fine-tuning (personalization) procedures, are provided in \prettyref{appendix:new_client_settings}. As demonstrated in \prettyref{table:new_client}, our method generalizes well even on clients with covariate shifts not encountered at training time. Moreover, observe that pFedFDA has the highest accuracy on the original clients, highlighting the efficacy of structured generative classifiers when less training data is available (i.e., having 50 rather than 100 clients).

\begin{table}[t]
\centering
\caption{\footnotesize
Evaluation of new-client generalization on CIFAR10 Dir(0.5).}
\begin{scriptsize}
\label{table:new_client}
\setlength{\tabcolsep}{3pt}
\resizebox{\textwidth}{!}{ %
\begin{tabular}{@{}ccccccccccccc@{}}
\toprule
\multirow{2}{*}{\textbf{}} & \multirow{2}{*}{\begin{tabular}[c]{@{}c@{}} \\ Original \\ Clients\end{tabular}} & \multicolumn{10}{c}{New Clients}                                                                                                                                                                                                                                                                                                                                                                                                                                     & \multicolumn{1}{l}{\textbf{}} \\ \cmidrule(l){3-13} 
                           &                                                                              & \begin{tabular}[c]{@{}c@{}}Clean\\ Data\end{tabular} & \begin{tabular}[c]{@{}c@{}}Motion\\ Blur\end{tabular} & \begin{tabular}[c]{@{}c@{}}Defocus\\ Blur\end{tabular} & \begin{tabular}[c]{@{}c@{}}Gauss\\ Noise\end{tabular} & \begin{tabular}[c]{@{}c@{}}Shot\\ Noise\end{tabular} & \begin{tabular}[c]{@{}c@{}}Impulse\\ Noise\end{tabular} & Frost              & Fog                & \begin{tabular}[c]{@{}c@{}}JPEG\\ Comp.\end{tabular} & Brightness         & Contrast                      \\ \midrule
FedAvg                     & .592(.07)                                                                    & .584(.08)                                            & .512(.09)                                             & .554(.08)                                              & .568(.07)                                             & .575(.07)                                            & .569(.07)                                               & .465(.08)          & .467(.08)          & .580(.07)                                            & .557(.08)          & .359(.10)                     \\
FedAvgFT                   & .716(.08)                                                                    & .709(.08)                                            & .689(.08)                                             & .704(.09)                                              & .695(.09)                                             & .699(.09)                                            & .696(.09)                                               & .680(.09)          & .672(.09)          & .711(.08)                                            & .707(.08)          & \underline{.688(.09)}               \\
FedBABU                    & .703(.10)                                                                    & .691(.09)                                            & .682(.08)                                             & .685(.09)                                              & .683(.09)                                             & .680(.09)                                            & .679(.09)                                               & .651(.10)          & .661(.09)          & .690(.08)                                            & .689(.09)          & .670(.09)                     \\
FedPAC                     & \underline{.727(.09)}                                                              & \underline{.724(.09)}                                      & \underline{.695(.09)}                                       & \underline{.708(.09)}                                        & \underline{.714(.09)}                                       & \underline{.712(.09)}                                      & \underline{.705(.09)}                                         & \underline{.682(.10)}    & \underline{.683(.09)} & \underline{.716(.09)}                                      & \underline{.718(.09)}    & .667(.09)                     \\
pFedFDA                    & \textbf{.738(.08)}                                                           & \textbf{.738(.08)}                                   & \textbf{.702(.09)}                                    & \textbf{.719(.09)}                                     & \textbf{.729(.08)}                                    & \textbf{.739(.07)}                                   & \textbf{.725(.08)}                                      & \textbf{.695(.09)} & \textbf{.684(.09)}    & \textbf{.738(.08)}                                   & \textbf{.733(.08)} & \textbf{.689(.09)}            \\ \bottomrule
\end{tabular}
}
\end{scriptsize}
\end{table}

\subsection{Ablation of Method Components}
We conduct two studies to verify the efficacy of our local-global interpolation method. In \prettyref{table:ablation}, we see that our interpolated estimates always perform better than using only local data, indicating the benefits of harnessing global knowledge. Learning separate $\beta$ terms for the means and covariance may be beneficial in low-sample or covariate-shift settings when the local distribution estimate may fluctuate further from the global estimate. However, using a single scalar $\beta$ appears sufficient and comes with the lowest computational cost (associated with the time to solve \prettyref{eq:beta}).
\label{results:ablation}
\begin{table}[!h]
\centering
\captionsetup{font=footnotesize}
\footnotesize
\caption{Ablation study on CIFAR100 with Dir(0.1) partition. \textbf{NB} denotes clients using only local data to estimate their feature distribution ($\beta_i=1$). \textbf{SB} denotes each client estimating a single $\beta_i$ for both the means and covariance, \textbf{MB} denotes clients computing $\beta_i$ terms for the means and covariance separately. We show the average computational overhead across all settings.}
\label{table:ablation}

\vspace{0.5em}
\setlength{\tabcolsep}{3pt}
\resizebox{\textwidth}{!}{ %
\begin{tabular}{@{}ccccccccccclc@{}}
\toprule
\multicolumn{3}{c}{$\beta$ Strategy}                                              &  & \multicolumn{3}{c}{Dir(0.1) Test Accuracy} &  & \multicolumn{3}{c}{Dir(0.5) Test Accuracy} & \multicolumn{1}{c}{} & Computation Overhead                  \\ \cmidrule(r){1-3} \cmidrule(lr){5-7} \cmidrule(lr){9-11} \cmidrule(l){13-13} 
NB                        & SB                        & MB                        &  & CIFAR100    & CIFAR100-25\%  & CIFAR100-S  &  & CIFAR100    & CIFAR100-25\%  & CIFAR100-S  &                      & (\% seconds/iter.)                \\ \midrule
\checkmark &                           &                           &  & .458(.08)   & .382(.09)      & .436(.08)   &  & .320(.06)   & .216(.05)      & .296(.06)   &                      & (0\%)                               \\
                          & \checkmark &                           &  & .523(.08)   & .396(.09)      & .487(.08)   &  & .385(.06)   & .266(.06)      & .361(.08)   &                      & $(\downarrow 22.35\%)$ \\
                          &                           & \checkmark &  & .514(.08)   & .423(.09)      & .480(.08)   &  & .379(.06)   & .275(.06)      & .373(.07)   &                      & $(\downarrow 36.11\%)$ \\ \bottomrule
\end{tabular}}
\end{table}

\begin{figure}[h!]
    \centering
    \includegraphics[width=0.8\textwidth]{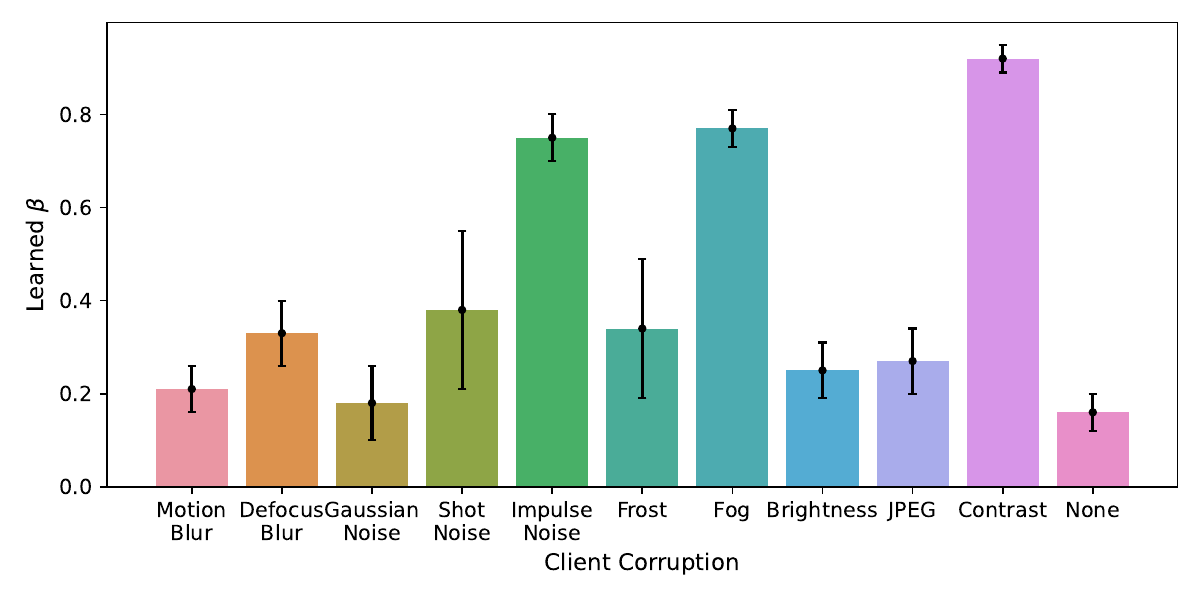}
    \caption{\footnotesize
    Comparison of client $\beta$ and local dataset corruption on CIFAR10-S.}
    \label{fig:learned_beta}
\end{figure}

We additionally visualize the spread of learned $\beta$ across clients as a function of their dataset corruption in \prettyref{fig:learned_beta}. As expected, clients with clean datasets rely more on global knowledge (smaller $\beta$ values) than corrupted clients. Moreover, corruptions with higher $\beta$ values (e.g., contrast) often align with the more difficult corruptions encountered in \prettyref{table:new_client}. 

\subsection{Communication and Computation}
\label{sec:computation}
The parameter count and relative communication load of our generative classifiers compared to a simple linear classifier varies depending on class count $C$ and feature %
dimension $d$. In our experimental configurations (datasets, architectures), the overhead in total parameter count ranges from 1.1\% to 6.8\%. See \prettyref{appendix:communication} for additional details.

In \prettyref{table:computation}, we compare the local training time (client-side computation) and total runtime of pFedFDA to baseline methods on CIFAR10. We observe a slight increase in training time relative to FedAvg, which can be attributed primarily to cost of learning our parameter interpolation coefficient $\beta$. However, this increase is comparable to the existing methods and is lower than representation-learning methods FedRep and FedPAC. This demonstrates the relative efficiency of our generative classifier formulation in comparison to classifiers obtained through local fine-tuning.
\begin{table}[!h]
\centering
\captionsetup{font=footnotesize}
\footnotesize
\caption{Comparison of system runtime on the CIFAR10 dataset. }
\label{table:computation}
\begin{tabular}{@{}lccccccc@{}}
\toprule
                     & FedAvg       & APFL         & Ditto        & FedBABU      & FedRep       & FedPAC       & pFedFDA      \\ \midrule
Local Training (sec) & 18.42 & 21.41 & 19.97 & 17.41 & 34.82 & 33.60 & 22.58 \\
Total Runtime (min)  & 61.41        & 71.35        & 66.87        & 58.11        & 116.95       & 146.3        & 77.71        \\ \bottomrule
\end{tabular}
\end{table}

\label{sec:experiments}

\section{Conclusion}
Balancing local model flexibility and generalization remains a central challenge in personalized federated learning (PFL). This paper introduces pFedFDA, a novel approach that addresses the bias-variance trade-off in client personalization through representation learning with generative classifiers. Our extensive evaluation on computer vision tasks demonstrates that pFedFDA significantly outper-forms current state-of-the-art methods in challenging settings characterized by covariate shift and data scarcity. Furthermore, our approach remains competitive in more general settings, showcasing its robustness and adaptability. The promising results underline the potential of our method to improve personalized model performance in real-world federated learning applications. Future work will focus on exploring the scalability of pFedFDA and its application to other domains.

\label{sec:conclusion}

\begin{ack}
We gratefully acknowledge the support from the National Science Foundation CAREER award under Grant No. 2340482, the Army Research Laboratory under Cooperative Agreement Number W911NF-23-2-0014, the Sony Faculty Innovation Award, and the National Defense \& Engineering Graduate (NDSEG) Fellowship Program. The views and conclusions contained in this document are those of the authors and should not be interpreted as representing the official policies, either expressed or implied,
of the Army Research Laboratory,
the National Science Foundation, 
or the U.S. government. The U.S. government is authorized to reproduce and distribute reprints for government purposes notwithstanding any copyright notation herein. We also thank Ming Xiang for valuable discussions and feedback on this work.
\end{ack}

\bibliographystyle{plain}
\bibliography{main}

\newpage
\appendix
\section{Limitations}
\label{appendix:limitations}
\begin{itemize}[leftmargin=*, noitemsep, nolistsep]
    \item 
    The selected class-conditional Gaussian distribution %
    may not work well for all neural network architectures. For example, if the output features are the result of an activation such as ReLU, a truncated Gaussian distribution may be a better model. Future work can look to exploit knowledge of the neural network architecture to improve the accuracy of the feature distribution estimate. 
    \item 
    In this work, 
    we leverage the insights from a fusion of global and local feature space. As in many applications there is often an underlying cluster structure between clients datasets, future works may explore the identification and efficient estimation of feature distributions of client clusters, in order to reduce the degree of bias introduced in client collaboration.
\end{itemize}

\section{Broader Impacts}
\label{appendix:impact}
Federated learning 
has become the main trend for distributed learning in recent years
and has deployed in many popular consumer devices such as Apple's Siri, Google's GBoard, and Amazon's Alexa.
Our paper addresses the practical limitations of personalization methods in adapting to clients with covariate shifts and/or limited local data, which is a central issue in cross-device FL applications. 
We are unaware of any potential negative social impacts of our work.

\section{Details of Experimental Setup}
All experiments are implemented in PyTorch 2.1 \cite{paszke2019pytorch} and were each trained with a single NVIDIA A100 GPU. Compute time per experiment ranges from approximately 2 hours for CIFAR10/100 and 20 hours for TinyImageNet.
Code for re-implementing our method is provided at the following GitHub URL: \url{https://github.com/cj-mclaughlin/pFedFDA}.
\label{appendix:numerical}

\subsection{Dataset Description}
\label{appendix:datasets}
The EMNIST \cite{cohen2017emnist} dataset contains over 730,000 28$\times$28 grayscale images of 62 classes of handwritten characters. The CIFAR10/CIFAR100 \cite{cifar} datasets contain 60,000 32$\times$32 color images in 10 and 100 different classes of natural images, respectively. TinyImageNet \cite{le2015tiny} contains 120,000 64$\times$64 color images of natural images.

\subsubsection{CIFAR-S Generation.}
\label{appendix:cifar-s}
We implement the following 10 common image corruptions at 5 levels of severity as described in \cite{hendrycks2019benchmarking}: Gaussian noise, shot (Poisson) noise, impulse noise, defocus blur, motion blur, fog, brightness, contrast, frost, JPEG compression. We apply a unique corruption-severity pair to all samples of the first 50 clients.

\subsubsection{Non-\iid Partitioning.}
\label{appendix:partition}
On CIFAR and TinyImageNet datasets, we distribute the proportion of samples of class $C$ across $M$ clients according to a Dirichlet distribution: $q_c, m \sim {\tt Dir}_{M}(\alpha)$, where we consider $\alpha \in (0.1, 0.5)$ as in \cite{lin2020ensemble}. 

We provide a visualization of Dirichlet partitioning strategies on CIFAR10 below. The size of each point represents the number of allocated samples. Notably, as $\alpha$ increases, Dir($\alpha$) becomes less heterogeneous. 
\begin{figure}[h]
\centering
\begin{subfigure}{.5\textwidth}
\centering
\includegraphics[width=\linewidth]{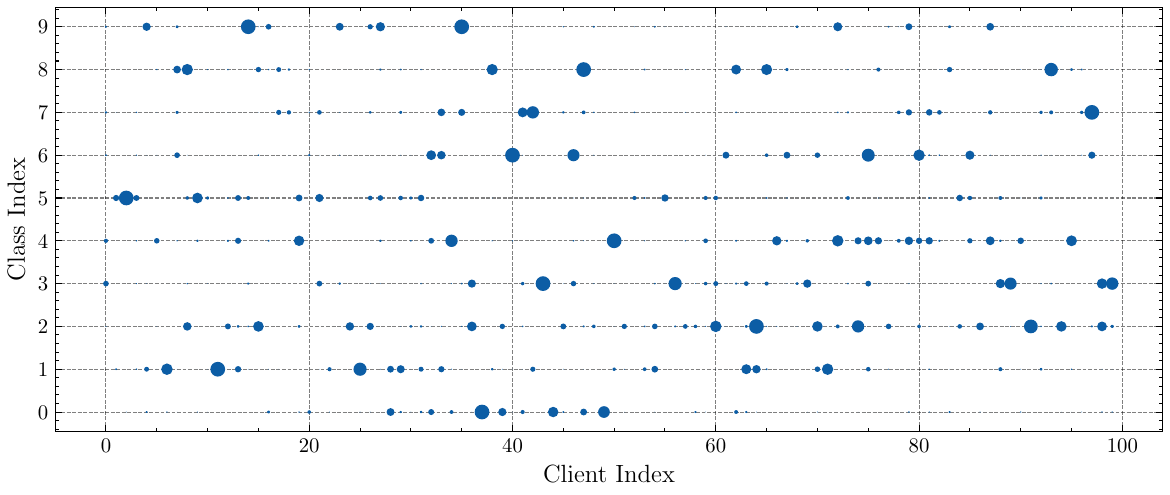}
\caption{CIFAR10 Dirichlet(0.1)}
  \label{fig:sub1}
\end{subfigure}%
\begin{subfigure}{.5\textwidth}
  \centering
    \includegraphics[width=\linewidth]{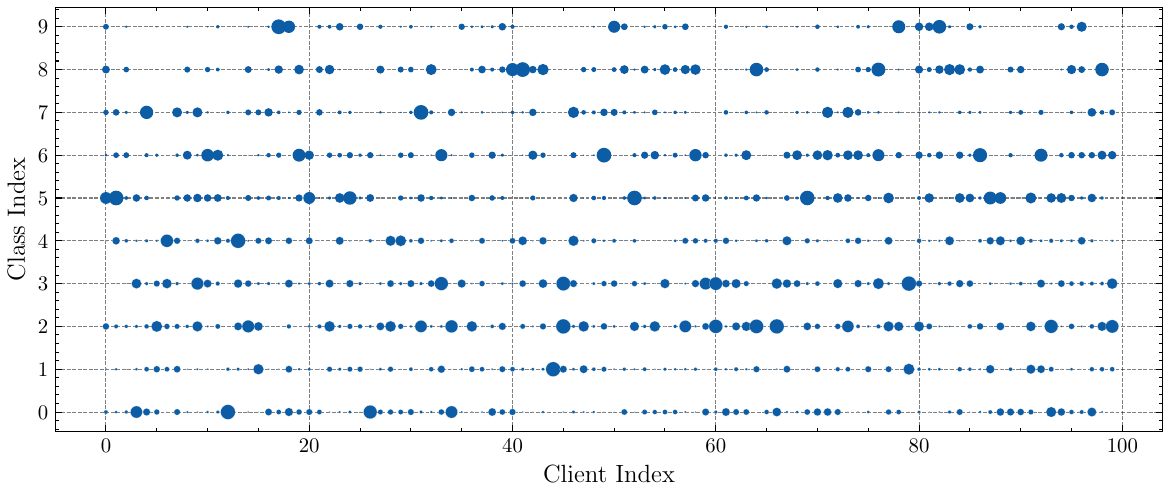}
\caption{CIFAR10 Dirichlet(0.5)}
\end{subfigure}
\caption{Comparison of Dirichlet Partitions on CIFAR10.}
\label{fig:local_epochs}
\end{figure}

\subsection{Training Settings}
\label{appendix:settings}
All methods are trained using mini-batch SGD for 200 global rounds with 5 local epochs of training. We use a fixed learning rate of 0.01, momentum of 0.5, and weight decay of 5e-4. The batch size is set to 50 for all experiments, except for EMNIST, where we use a batch size of 16. We sample the set of active clients uniformly with probability q=0.3 for CIFAR and TinyImageNet and q=0.03 for EMNIST. The last global round of training employs full client participation. We split the data of each client 80-20\% between training and testing. 

\noindent
\textbf{Hyper-parameters.} For APFL, we tune $\alpha$ over [0.25, 0.5, 0.75, 1.0], and set $\alpha=0.25$. For pFedMe, we tune $\lambda$ over [1.0, 5.0, 10.0, 15.0] and set $\lambda=5.0$. For Ditto, we use five local epochs for personalization and tune $\mu$ over [0.05, 0.1, 0.5, 1.0, 2.0] and set $\mu=1.0$. For FedRep and FedBABU, we use five local epochs for training the head parameters. For FedPAC, we tune $\lambda$ over [0.1, 0.5, 1.0, 5.0, 10.0], and set $\lambda=1.0$. FedPAC uses one local epoch for training head parameters with a higher learning rate of 0.1, following the original implementation.

\noindent 
\subsection{Evaluation on New Clients}
\label{appendix:new_client_settings}
Our fine-tuning procedure on new clients largely follows the methodology above. For FedAvgFT, we fine-tune the global model for five local epochs. For FedBABU and FedPAC, we personalize the model in 2 different ways and report the best result: (1) fine-tuning only the head for 5 local epochs, and (2) fine-tuning both the body and head for 5 local epochs. For pFedFDA, each new client estimates their local interpolated statistics (i.e., lines \ref{line:gauss_estimate}-\ref{line:interpolation} of \Cref{algorithm:pfedfda_pseudocode}) to obtain a personalized generative classifier. 

For our covariate shift evaluation, we apply a medium severity corruption (level 3) to all samples. 

\section{Additional Results}
\subsection{Multi-Domain FL}
In \Cref{table:digit5}, we present results on the DIGIT-5 domain generalization benchmark \cite{zhou2020learning}. This presents an alternate form of covariate shift, as the data from each client is drawn from one of 5 datasets (SVHN, USPS,
SynthDigits, MNIST-M, and MNIST). In particular, we use 20 clients trained with full participation, and assign 4 clients to each domain. Within each domain, we use the Dirichlet(0.5) partitioning strategy to assign data to each client. We observe that pFedFDA is effective in all settings, but has the most significant benefits over prior work in the low-data regime. 
\begin{table}[h]
\centering
\captionsetup{font=footnotesize}
\caption{\footnotesize Results on multi-domain DIGIT-5 benchmark for varying data volumes. }
\label{table:generalization}
\vspace*{\baselineskip}
\begin{small}
\begin{tabular}{@{}lccccc@{}}
\toprule
DIGIT-5 \% Samples & 25                           & 50                           & 75                           & 100                          & Avg. Improvement \\ \midrule
Local              & 76.84                 & 83.11                  & 86.97                 & 88.51                  & -                \\
FedAvg             & 81.75 (+4.91)         & 85.09 (+1.98)          & 87.41 (+0.44)          & 88.19 (+0.32)          & 1.91             \\
FedAvgFT           & \underline{85.61 (+8.77)} & \underline{88.72 (+5.61)} & 90.75 (+3.78)          & \textbf{91.73 (+3.22)} & \underline{5.34}    \\
Ditto              & 83.85 (+7.01)          & 85.53 (+2.42)          & 87.43 (+0.46)          & 88.80 (+0.29)          & 2.54             \\
FedPAC             & 82.78 (+5.94)          & 87.94 (+4.83)          & \textbf{91.12 (+4.15)} & 91.04 (+2.53)          & 4.36             \\
pFedFDA            & \textbf{86.54 (+9.70)} & \textbf{90.05 (+6.94)} & \underline{90.75 (+3.78)} & \underline{91.56 (+3.05)}  & \textbf{5.86}    \\ \bottomrule
\end{tabular}
\end{small}
\label{table:digit5}
\end{table}

\subsection{Effect of Local Epochs}
In many FL settings, we would like clients to perform more local training between rounds to reduce communication costs. However, too much local training can cause the model to diverge. In \Cref{fig:local_epochs}, we compare the effect of the local amount of epochs for CIFAR100 and CIFAR100-S-25\% sample datasets. We observe that (1) pFedFDA outperforms FedAvgFT at all equivalent budgets of $E$, (2) both methods follow exhibit a general plateau in accuracy after $E=5$, and (3) pFedFDA learns much faster than FedAvgFT, with significantly higher accuracy for $E=1$.

\begin{figure}[h]
\centering
  \centering
  \includegraphics[width=0.7\linewidth]{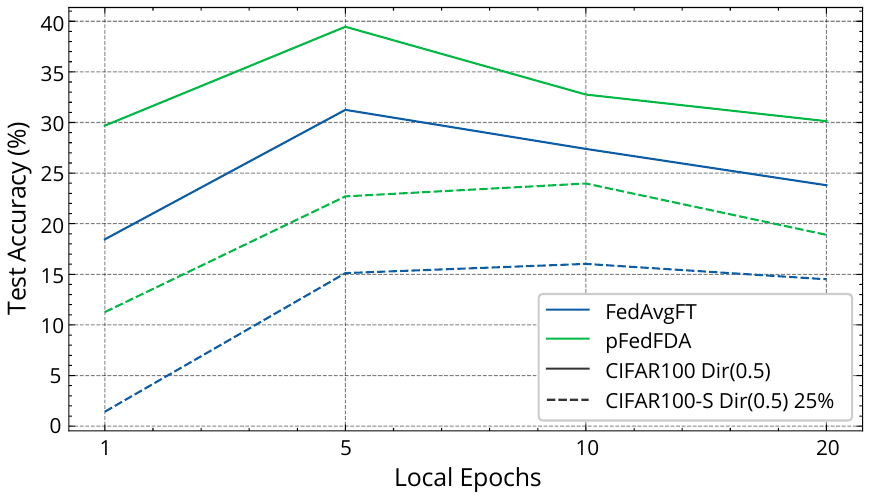}
  \caption{Comparison of average test accuracy with varying local epochs on CIFAR100.}
\label{fig:local_epochs}
\end{figure}

\subsection{Communication Load Examples}
In \Cref{table:communication}, we compare the number of distinct parameters in our Gaussian estimates to that of a typical linear classifier for the models and datasets used in this paper, along with some additional examples. We display the resulting overhead relative to the base parameter count of the shared representation backbone.
\begin{table}[h!]
\centering
\captionsetup{font=footnotesize}
\footnotesize
\caption{Comparison of communication load (parameters/iter.) between our Gaussian distribution parameters ($\mu, \Sigma$) and standard linear classifiers. }
\label{table:communication}
\begin{tabular}{@{}lcccc@{}}
\toprule
Parameters                                                                           & Backbone & \begin{tabular}[c]{@{}c@{}}Linear Classifier \\ $(C \times (d+1))$ \end{tabular} & \begin{tabular}[c]{@{}c@{}} Gaussian ($\bm{\mu}$, $\Sigma$) \\ $(C \times d + \frac{1}{2}(d^{2}+d))$ \end{tabular} & $\Delta$ Overhead  \\ \midrule
\begin{tabular}[c]{@{}l@{}}EMNIST-CNN\\  (EMNIST, $C=62$)\end{tabular}                  & 115776                  & 7998                                                                                               & 16192                                                                                                              & 6.620\%                     \\      
\begin{tabular}[c]{@{}l@{}}CIFAR-CNN\\  (CIFAR100, $C=100$)\end{tabular}                  & 106400                  & 12900                                                                                               & 21056                                                                                                              & 6.837\%                     \\            
\begin{tabular}[c]{@{}l@{}}ResNet18\\ (TinyImageNet, $C=200$)\end{tabular}       & 11167212                & 102600                                                                                              & 233728                                                                                                             & 1.164\%                     \\
\begin{tabular}[c]{@{}l@{}}MobileNetV3-Small\\ (ImageNet, $C=1000$)\end{tabular} & 927008                  & 1615848                                                                                             & 1548800                                                                                                            & -2.637\%                     \\ \bottomrule
\end{tabular}
\end{table}
\label{appendix:communication}

\subsection{Runtime of Method Components}
In \Cref{table:component_runtime}, we evaluate the proportion of each local iteration of pFedFDA associated with each line of our algorithm. \textbf{Network Passes} refers to the time taken to train the base network parameters $\phi$ (Line 7 of Alg. 1). \textbf{Mean/Covariance Est.} refers to the time taken to estimate the local mean and covariance from features extracted during model training (Line 8 of Alg. 1). \textbf{Interpolation Optimization} refers to the time taken to optimize the local coefficient $\beta$ (Line 9 of Alg. 1). Overall, we find that the majority of the overhead of our method comes from estimating the interpolation parameter $\beta$.

\begin{table}[h]
\centering
\caption{\footnotesize Percentage (average (std)) of the local training time associated each component of our algorithm. }
\footnotesize
\vspace*{\baselineskip}
\begin{small}
\begin{tabular}{@{}cccc@{}}
\toprule
             & Network Passes & Mean/Covariance Est. & Interpolation Optimization \\ \midrule
CIFAR10      & 84.88 (6.79)\%  & 0.765 (0.281)\%       & 14.36 (6.62)\%               \\
CIFAR100     & 77.70 (5.75)\%  & 2.861 (0.899)\%        & 19.43 (5.70)\%               \\
TinyImageNet & 87.41 (1.50)\%   & 2.701 (0.659)\%        & 9.886 (1.14)\%               \\ \bottomrule
\end{tabular}
\label{table:component_runtime}
\end{small}
\end{table}

\newpage
\section{On the Bias-Variance Tradeoff}
\label{appendix:proof}
This section justifies the bias-variance tradeoff under some simplified technical assumptions. 
For simplicity, we assume that at any given round, the extracted feature vectors for a class are independent. %
We illustrate the bias-variance tradeoff in estimating the mean feature of a given class $c$ at round $t$. 
\begin{proof}[\bf Proof of Theorem \ref{Theorem:bias-variance}]
For ease of exposition, we drop the time index and class index. 

Let $i$ be an arbitrary client with local dataset size $n_i$ of class $c$. Let $N$ be the total data volume of class $c$ over the entire FL system. Assuming that the distribution of client $i$'s features $z$ follow a multivariate Gaussian distribution $\calN(\theta_i, \Sigma_i)$, and the global feature distribution follows $\calN(\theta, \Sigma)$ where $\theta_g := \sum_{i=1}^M n_i \theta_i / (\sum_{i\in[M]}n_i)$, $\Sigma_g := \sum_{i=1}^M n_i^2 \Sigma_i /(\sum_{i\in[M]}n_i)^2$. Note $\theta_i, \theta_g$ are deterministic parameters.

We denote the local and global mean estimates as:

\begin{align*}
 \mu_i := \frac{1}{n_i}\sum_{j=1}^{n_i} z_i^j, ~~ \text{and} ~~ \mu_g := \frac{1}{N}\sum_{i=1}^M\sum_{j=1}^{n_i} z_i^j.    
\end{align*}

Let $\hat{\mu}_i$ be the local estimate that interpolates between local and global knowledge, defined as 
\begin{align}
\label{eq: final local estimate}    
\hat{\mu}_i : = \beta \mu_i + (1-\beta) \mu_g. %
\end{align}

We will focus on bounding the high probability local estimation error $\norm{\hat{\mu_i} - \expect{\mu_i}}$.

Note that Eq.\eqref{eq: final local estimate} can be further expanded as 
\begin{align*}
\hat{\mu}_i &= \beta \mu_i + (1-\beta) \pth{\frac{n_i}{N} \mu_i + \sum_{i^{\prime} \not= i} \frac{n_{i^{\prime}}}{N} \mu_{i^{\prime}}} \\
& = (\beta + (1-\beta)\frac{n_i}{N})\mu_i + (1-\beta)\sum_{i^{\prime} \not= i} \mu_{i^{\prime}}\\
& = \gamma \mu_i + (1-\beta) \Bar{\mu}, 
\end{align*}
where $\gamma := \beta + (1-\beta)\frac{n_i}{N}$, and $\Bar{\mu} := \frac{1}{N}\sum_{i^{\prime} \not= i} \sum_{j=1}^{n_{i^{\prime}}}\mu_{i^{\prime}}^j$. 

Thus $\mu_i \sim \calN(\theta_i, \frac{1}{n_i}\Sigma_i)$ and $\Bar{\mu} \sim \calN(\frac{N\theta_g - n_i\theta_i}{N}, \frac{N\Sigma_g - n_i \Sigma_i}{N^2})$. Since $\mu_i$ and $\Bar{\mu}$ are independent, we have

\begin{align*}
\hat{\mu}_i - \theta_i ~ \sim ~ \calN\pth{(1-\beta)(\theta_g-\theta_i), \,\, 
\gamma^2 \frac{1}{n_i}\Sigma_i + (1-\beta)^2 \frac{N\Sigma_g - n_i\Sigma_i}{N^2}}.    
\end{align*}

Let $\bm{g} \sim \calN(\bm{0}, \identity)$. We have 

\begin{align*}
\hat{\mu}_i - \theta_i = (1-\beta)(\theta_g-\theta_i) + \Bar{\Sigma}^{1/2} \bm{g},
\end{align*}

where $\hat{\Sigma}^{1/2}$ is the square root matrix of $\hat{\Sigma} := \gamma^2 \frac{1}{n_i}\Sigma_i + (1-\beta)^2 \frac{N\Sigma_g - n_i\Sigma_i}{N^2}$. It holds that
\begin{align*}
\norm{\hat{\mu}_i - \theta_i}^2 
&= (1-\beta)^2 \norm{\theta_g - \theta_i}^2 + 2(1-\beta) \iprod{\theta_g - \theta_i}{\hat{\Sigma}^{1/2} \bm{g}} \\
& \qquad + \iprod{\hat{\Sigma}^{1/2} \bm{g}}{\hat{\Sigma}^{1/2} \bm{g}}. 
\end{align*}

Taking the expectation with respect to the randomness in the Gaussian random variable $\bm{g}$ and by the law of total expectation, we have
\begin{align*}
\expect{2(1-\beta) \iprod{\theta_g - \theta_i}{\hat{\Sigma}^{1/2} \bm{g}}} 
= 2(1-\beta) \iprod{\theta_g - \theta_i}{\hat{\Sigma}^{1/2} \expect{\bm{g}}} = 0,
\end{align*}

and 
\begin{align*}
\expect{\iprod{\hat{\Sigma}^{1/2} \bm{g}}{\hat{\Sigma}^{1/2} \bm{g}}} 
&\overset{(a)}{=} \expect{\bm{g}^{\top} \hat{\Sigma} \bm{g}} \\ 
&= \expect{\bm{g}^{\top} \gamma^2 \frac{1}{n_i}\Sigma_i\bm{g}} 
+ \expect{\bm{g}^{\top} (1-\beta)^2 \frac{N\Sigma_g - n_i\Sigma_i}{N^2} \bm{g}} \\
& = \gamma^2 \frac{1}{n_i} \Tr(\Sigma_i) + (1-\beta)^2 \frac{N\Tr(\Sigma_g) - n_i\Tr(\Sigma_i)}{N^2}. 
\end{align*}

where equality $(a)$ holds because $(\hat{\Sigma}^{1/2})^\top (\hat{\Sigma}^{1/2}) = \hat{\Sigma}$ as $\hat{\Sigma}^{1/2}$ is symmetric.

By Hanson-Wright inequality \cite[Theorem 6.2]{vershynin2018high}, we conclude that with probability at least $1-\delta$ (for any given $\delta\in (0,1)$), 

\begin{align*}
\norm{\hat{\mu}_i - \theta_i}^2 
 &\le 
 (1-\beta)^2 \norm{\theta_g - \theta_i}^2  \\
& ~~~
+ (\frac{\beta^2}{n_i} + \frac{2\beta(1-\beta)}{N})\Tr(\Sigma_i) + (1-\beta)^2 \frac{1}{N}\Tr(\Sigma_g) \\
& ~~~
+ 4\lnorm{(\frac{\beta^2}{n_i} + \frac{2\beta(1-\beta)}{N})\Tr(\Sigma_i) + (1-\beta)^2 \frac{1}{N}\Tr(\Sigma_g)}{\rm F} \max\sth{\sqrt{\frac{\log 1/\delta}{c}}, \frac{\log 1/\delta}{c}} \\
&\overset{(b)}{\le} (1-\beta)^2 \norm{\theta_g - \theta_i}^2 \\
& ~~~
+ \frac{\beta^2 + 2 \beta (1-\beta)}{n_i} 
\Tr(\Sigma_i) + \frac{(1-\beta)^2}{N}\Tr(\Sigma_g) \\
& ~~~
+ 4\lnorm{(\frac{\beta^2}{n_i} + \frac{1}{2})\Tr(\Sigma_i) + (1-\beta)^2 \frac{1}{N}\Tr(\Sigma_g)}{\rm F} \max\sth{\sqrt{\frac{\log 1/\delta}{c}}, \frac{\log 1/\delta}{c}} \\
&\overset{(c)}{\le} (1-\beta)^2 \norm{\theta_g - \theta_i}^2 \\
& ~~~
+ \frac{2 \beta -\beta^2}{n_i}
\Tr(\Sigma_i) + \frac{(1-\beta)^2}{N}\Tr(\Sigma_g) \\
& ~~~
+ 4\pth{\sqrt{\frac{\log 1/\delta}{c}} + \frac{\log 1/\delta}{c}} \pth{
\frac{2 \beta -\beta^2}{n_i} 
\sqrt{\Tr(\Sigma_i^2)} + 
\frac{(1-\beta)^2}{N} \sqrt{\Tr(\Sigma_g^2)}}  \\
&\overset{(d)}{\le} (1-\beta)^2 \norm{\theta_g - \theta_i}^2 \\
& ~~~
+ \qth{1 + 4\pth{\sqrt{\frac{\log 1/\delta}{c}}+\frac{\log 1/\delta}{c}}}
\pth{
\frac{2\beta}{n_i}
\Tr(\Sigma_i) + 
\frac{(1-\beta)^2}{N}\Tr(\Sigma_g)},
\end{align*}
where $c>0$ is some absolute constant,
inequality $(b)$ holds as $a \cdot b \le (\frac{a+b}{2})^2$, and $N \ge n_i$, %
inequality $(c)$ holds because of triangular inequality $\fnorm{A + B} \le \fnorm{A} + \fnorm{B}$,
that $\fnorm{A} = \sqrt{\fnorm{A}^2} = \sqrt{\Tr{(A^\top A)}} = \sqrt{\Tr(A^2)}$ if matrix $A$ is symmetric,
and that $\max\{a,b\} \le a+b$,
inequality $(d)$ holds because $\Tr(A^2) \le (\Tr(A))^2$ for positive semidefinite matrix $A$ and that $\Tr(\Sigma_i)$,
$\beta^2 \ge 0$,
and $\Tr(\Sigma_g)$ are by definition non-negative.

\end{proof}

The first term $(1-\beta)^2\norm{\theta_g - \theta_i}^2$ is the bias introduced when client $i$ uses global knowledge; the smaller the $\beta$, the more bias introduced. The last term reveals the interaction of $\beta$ and the tradeoff between local and global variance. When $\beta$ approaches 0, we have the global feature variance $\Tr(\Sigma)$ reduced by the average of $N$ global samples. When $\beta$ approaches 1, we have local feature variance $\Tr(\Sigma_i)$ reduced by the average of only $n_i$ local data. Thus the bias-variance tradeoff on client $i$ crucially depends on the degree of local-global distribution shift, $\norm{\theta_g - \theta_i}^2$, the local data volume $n_i$ and its quality (i.e., $\Sigma_i$), and the volume and quality of the data across clients $N, \Sigma_g$.

\end{document}